\documentclass[12pt]{article}

\usepackage{newtxtext,newtxmath}
\usepackage{txfonts}
\usepackage{pifont}
\usepackage{graphicx}
\usepackage[letterpaper,margin=1in]{geometry}

\renewenvironment{abstract}
	{\quotation}
	{\endquotation}

\date{}

\makeatletter
\renewcommand{\fnum@figure}{\textbf{Figure \thefigure}}
\renewcommand{\fnum@table}{\textbf{Table \thetable}}
\makeatother

\usepackage{scicite}

 \usepackage{url}
\definecolor{myblue}{RGB}{0, 32, 91}

\usepackage[colorlinks=true, urlcolor=myblue]{hyperref} 

\def\scititle{
	\textsc{ManipulationNet}: An Infrastructure for Benchmarking Real-World Robot Manipulation \\with Physical Skill Challenges and Embodied Multimodal Reasoning
}
\title{\bfseries \boldmath \scititle}

\author{
	Yiting Chen$^{1}$,
	Kenneth Kimble$^{2}$,
        Edward H. Adelson$^{3}$, 
        Tamim Asfour$^{4}$, \and
        Podshara Chanrungmaneekul$^{1}$,
        Sachin Chitta$^{5}$,
        Yash Chitambar$^{6}$,
        Ziyang Chen$^{6}$,\and
        Ken Goldberg$^{6}$,
        Danica Kragic$^{7}$, 
        Hui Li$^{5}$,
        Xiang Li$^{8}$, 
        Yunzhu Li$^{9}$,
        Aaron Prather$^{10}$,\and
        Nancy Pollard$^{11}$, 
        Maximo A. Roa-Garzon$^{12}$, 
        Robert Seney$^{2}$, 
        Shuo Sha$^{9}$, \and
        Shihefeng Wang$^{8}$, 
        Yu Xiang$^{13}$,
        Kaifeng Zhang$^{9}$,
        Yuke Zhu$^{14, 15}$,
	Kaiyu Hang$^{1}$\\
        \href{https://manipulation-net.org}{\small\textbf{\texttt{https://manipulation-net.org}}} \\
     \small$^1$Rice University $^2$U.S. National Institute of Standards and Technology \\ \small$^3$Massachusetts Institute of Technology $^4$Karlsruhe Institute of Technology \\\small$^5$Autodesk Research $^6$University of California, Berkeley $^7$KTH Royal Institute of Technology \\\small$^8$Tsinghua University $^9$Columbia University $^{10}$ASTM International \\\small$^{11}$Carnegie Mellon University $^{12}$German Aerospace Center (DLR) \\\small$^{13}$University of Texas at Dallas $^{14}$University of Texas at Austin $^{15}$NVIDIA Research
}

\begin{document} 

\maketitle

\begin{abstract} \bfseries \boldmath
\begin{itemize}
Dexterous manipulation enables robots to purposefully alter the physical world, transforming them from passive observers into active agents in unstructured environments. This capability is the cornerstone of physical artificial intelligence. Despite decades of advances in hardware, perception, control, and learning, progress toward general manipulation systems remains fragmented due to the absence of widely adopted standard benchmarks. The central challenge lies in reconciling the variability of the real world with the reproducibility and authenticity required for rigorous scientific evaluation. To address this, we introduce \emph{ManipulationNet}, a global infrastructure that hosts real-world benchmark tasks for robotic manipulation. ManipulationNet delivers reproducible task setups through standardized hardware kits, and enables distributed performance evaluation via a unified software client that delivers real-time task instructions and collects benchmarking results. As a persistent and scalable infrastructure, ManipulationNet organizes benchmark tasks into two complementary tracks: 1) the Physical Skills Track, which evaluates low-level physical interaction skills, and 2) the Embodied Reasoning Track, which tests high-level reasoning and multimodal grounding abilities. This design fosters the systematic growth of an interconnected network of real-world abilities and skills, paving the path toward general robotic manipulation.   By enabling comparable manipulation research in the real world at scale, this infrastructure establishes a sustainable foundation for measuring long-term scientific progress and identifying capabilities ready for real-world deployment.

\end{itemize}
\end{abstract}

\noindent
\section{Introduction}
Robotic manipulation refers to the physical processes through which robots interact with objects in their environment to achieve specified goals~\cite{mason2001mechanics}. This capability comprises a wide range of fundamental skills~\cite{mason2018toward}, including but not limited to grasping, releasing, insertion, pushing, and pulling, which are essential across domains such as manufacturing, logistics, and healthcare. While being fundamentally physical, manipulation also requires cognitive abilities~\cite{billard2025roadmap} such as perception, reasoning, and multimodal grounding, which enable robots to interpret their environment, understand and contexualize task requirements and constraints, and generalize physical skills to novel situations. However, despite decades of endeavors, dexterous manipulation remains largely confined to structured environments~\cite{billard2019trends}, with robots continuing to struggle in unstructured and dynamic settings. According to the International Federation of Robotics (IFR), although more than 4.3 million industrial robots operate worldwide~\cite{IFR2024Industrial}, their success is concentrated in controlled factory settings, while service robots, despite a 30\% sales increase in 2023~\cite{IFR2024Service}, remain limited to tasks such as delivery and transport that avoid complex, contact-rich manipulation. Unlocking robust manipulation capabilities is therefore the critical next step in realizing the projected growth of the robot market and in enabling robots to effectively improve human lives. Central to this progress is the role of benchmarks.

Benchmarks transform isolated research advances into cumulative scientific progresses. Landmark examples such as \textit{ImageNet}~\cite{deng2009imagenet} in computer vision and \textit{GLUE}~\cite{wang2018glue} in natural language processing have reshaped their respective fields by providing standard evaluation criteria and driving collective innovation toward shared goals. Within the broader landscape of artificial intelligence, robotics stands as a domain poised to benefit from large-scale and standardized benchmarking efforts. However, unlike fields that operate on static datasets, benchmarking robotic manipulation requires dynamic interactions with the physical world, leading to substantial heterogeneity across tasks, objects, and contexts. To address this heterogeneity, manipulation benchmarks should provide standardized task setups and directly evaluate the system's performance through manipulation metrics.

Existing efforts can be broadly grouped into simulation-based and real-world evaluations. Simulation-based benchmarks offer reproducible tasks and accessibility at scale, but their imperfect approximations of contact dynamics render results incapable of fully reflecting the true manipulation capabilities of robotic systems. Real-world evaluation offers physical fidelity and often requires centralized setups to ensure task consistency, which is difficult to scale due to limited accessibility. In the context of manipulation benchmarking, we define task \emph{authenticity} as requiring participants to perform the same task with identical objects under specified protocols, with reported results faithfully reflecting the execution outcomes. \emph{Realism} refers to the real-world evaluation, and  \emph{accessibility} to broad community participation. Viewed in this way, this persistent imbalance between realism, authenticity, and accessibility has prevented the emergence of a widely adopted benchmark for robotic manipulation. As a result, progress remains fragmented even for ostensibly identical tasks.

To break this impasse and unlock field-wide progress, we introduce ManipulationNet, a global, community-governed initiative that balances realism, authenticity, and accessibility to enable benchmarking real-world robot manipulation at scale. 
ManipulationNet provides a persistent and scalable infrastructure to support comparable manipulation research across the globe. It integrates a hybrid centralized–decentralized architecture that 1) distributes standardized object sets and protocols worldwide to support reproducible task setups, 2) enables distributed performance collection through local submission clients to a centralized server while preserving task authenticity, and 3) provides centralized performance evaluation to ensure comparability of the results. In this way, researchers can benchmark customized manipulation systems on shared tasks at any time and from any location.
For clarity, we denote the local submission client as \texttt{mnet-client} and the centralized server as \texttt{mnet-server} throughout this paper.

Tasks hosted by ManipulationNet are diagnostic, each targeting a specific skill or ability. The framework organizes these tasks into two complementary tracks. The Physical Skills Track evaluates manipulation from a physical interaction perspective, assessing how well robots execute robust sensorimotor skills under real-world physical constraints. The Embodied Reasoning Track minimizes the physical burden but emphasizes reasoning and multimodal grounding, assessing how robots contexualize and translate natural language instructions and visual inputs into applicable actions for manipulation. By integrating tasks across both physical and cognitive domains, ManipulationNet constructs an interconnected network that reveals how physical skills can be composed into more complex behaviors, how cognitive abilities can be grounded in effective and reliable interactions, and how knowledge can transfer across tasks and contexts.



This interconnected design paves the path toward general robotic manipulation.
In the near term, ManipulationNet aims to drive collective efforts around common benchmarks; in the medium term, it will expand to cover a broader spectrum of manipulation challenges; and in the long term, it will establish sustainable records of robot capabilities. Such a record will not only identify systems that are ready for real-world deployment, but also chart the trajectories of field-wide progresses, reveal persistent capability gaps, and guide research priorities for the field.   By providing a transparent and cumulative account of what robots can and cannot yet do, ManipulationNet lays the foundation for both scientific discovery and trustworthy adoption in real-world settings. 


\subsection{A longstanding need and effort}

Benchmarking manipulation research as an open problem has been attracting attention for four decades~\cite{del2006benchmarks, collins1985development}, which primarily involves systematically evaluating a robotic system’s performance by assessing its manipulation outcomes under a well-defined task setup.  Prior initiatives in benchmarking manipulation can be broadly grouped into three categories: standardized object sets (often paired with task protocols), competitions, and simulation-based benchmarks. Together, these complementary approaches have shaped the field and laid the foundation for subsequent advances.

\textbf{Object sets and protocols} are the fundamental elements involved in manipulation, and their standardization is a critical step towards reproducible and comparable research. The Columbia grasp database~\cite{goldfeder2009columbia} pioneered this direction by introducing a repository of synthetic object models. These models were reused from the Princeton Shape Benchmark~\cite{shilane2004princeton}, annotated with grasp candidates and quality metrics for simulation-based evaluation~\cite{miller2004graspit}. By fixing the virtual object set, this effort laid the foundation for systematic comparison of grasping algorithms in simulation. To narrow the gap to the physical world, the Karlsruhe Institute of Technology (KIT) object models database~\cite{kasper2012kit} provided accurate 3D scans of everyday objects. Building on this idea, the Yale-CMU-Berkeley (YCB) Object and Model Set~\cite{calli2015ycb} paired physical items with digital counterparts in a distributed kit, supporting manipulation research in both simulation and real-world settings.The YCB Object and Model Set has since been extended with diverse task protocols, including hand design~\cite{coulson2021elliott, llop2019anthropomorphic}, bimanual manipulation~\cite{garcia2020benchmarking, chatzilygeroudis2020benchmark}, grasp resilience~\cite{negrello2020benchmarking} and planning~\cite{bottarel2020graspa, bekiroglu2019benchmarking}, in-hand manipulation~\cite{cruciani2020benchmarking}, pick-and-place~\cite{morgan2019benchmarking, khargonkar2024scenereplica, bottarel2020graspa}, aerial manipulation~\cite{suarez2020benchmarks}, picking-in-clutter~\cite{leitner2017acrv}, Rubik's cube manipulation~\cite{yang2020benchmarking} and assembly tasks~\cite{calli2015benchmarking}. The National Institute of Standards and Technology (NIST) further advanced physical benchmarking~\cite{kimble2020benchmarking} through standardized assembly task boards (ATBs) and protocols, enabling comparable research of contact-rich manufacturing-based skills. 
Related efforts have also introduced object sets for task-specific evaluation, such as 6-DoF pose estimation~\cite{tyree20226}, functional manipulation~\cite{luo2025fmb} and furniture assembly~\cite{heo2023furniturebench}. Collectively, standardized object sets and protocols permit objective assessment across different systems.

\textbf{Real-world competitions} have long functioned as de facto benchmarks by providing common tasks and centralized evaluation infrastructures. RoboCup~\cite{kitano1997robocup}, launched in 1997, established a model of year-over-year iterative competition and later expanded to include RoboCup@Home~\cite{wisspeintner2009robocup} for domestic service, RoboCup@Work~\cite{kraetzschmar2014robocup} for logistics, and RoboCup Rescue Robot League~\cite{Kitano1999RoboCupRescue} for disaster scenarios where manipulation is central. Industry-driven efforts such as the Amazon Picking Challenge (APC)~\cite{correll2016analysis} pushed the field toward warehouse-style object picking and stowing. Publicly funded initiatives, most notably the DARPA Robotics Challenge~\cite{krotkov2018darpa}, featured manipulation tasks central to disaster-response scenarios, including vehicle driving, valve turning, tool operation, and material handling. The Robot Competitions Kick Innovation In Cognitive Systems and Robotics (RoCKIn) project~\cite{amigoni2013benchmarking} explicitly linked competitions to structured benchmarking through standardized tasks in domestic~\cite{schneider2014rockin} and industrial~\cite{dwiputra2014rockin} scenarios for reproducible scientific evaluation. Similarly, the European Robotic Challenge (EuRoC)~\cite{siciliano2014euroc} combined simulation-based qualification with field trials in real industrial environments, and other notable competitions, such as the World Robot Summit (WRS)~\cite{Tanaka2020WRS}, have also driven advancements in robotics and manipulation. Within academic communities, specialized competitions~\cite{sun2021research} have been organized, such as the Robotic Grasping and Manipulation Competition (RGMC)~\cite{sun2024robotic}, to evaluate a wide range of grasping and manipulation tasks that reflect service and industry scenarios. The Real Robot Challenge~\cite{bauer2022real}, Train Offline Test Online (TOTO) Benchmark~\cite{zhou2023train} and Open Cloud Robot Table Organization Challenge (OCRTOC)~\cite{liu2021ocrtoc} provide remote access to identical robotic platforms for centralized policy evaluation. These events validate systems under practical conditions, highlighting robustness, generalization, and applicability in the real world.

\textbf{Simulation-based benchmarks} enable scalable evaluation under tightly controlled conditions, relying on physics engines~\cite{physx, diankov2008openrave, coumans2015bullet, todorov2012mujoco} to model perception, kinematics, and dynamics. As an early milestone, OpenGRASP~\cite{ulbrich2011opengrasp} introduces holistic environments for grasping and dexterous manipulation, establishing the feasibility of virtual evaluation. VisGraB~\cite{kootstra2012visgrab} presents a simulation toolbox to evaluate vision-based grasp planners for unknown objects grasping in unstructured environments. With the rise of robot learning, SURREAL~\cite{fan2018surreal} is developed as a scalable infrastructure for distributed reinforcement learning (RL), providing an accessible set of benchmarking tasks to support reproducible robot manipulation research. Initiated from SURREAL, RoboSuite~\cite{zhu2020robosuite} advances the field by providing a modular framework that supports flexible task construction and reproducible environments for robot learning research. In parallel, task diversity expanded through benchmarks such as Meta-World~\cite{yu2020meta}, which introduces dozens of distinct tasks for multi-task and meta RL, and RLBench~\cite{james2020rlbench}, which emphasizes vision-guided manipulation with tasks of varying difficulty and extensive demonstration data for imitation learning. Subsequent efforts shifted toward richer embodiment and broader generalization. BEHAVIOR~\cite{srivastava2022behavior, li2023behavior} targets household activities that require long-range, goal-directed execution in realistic environments, while ManiSkill~\cite{mu2maniskill, gumaniskill2} stresses geometric diversity by incorporating large intraclass variations in object shape and topology. Beyond physical interaction, Calvin~\cite{mees2022calvin} and ALFRED~\cite{shridhar2020alfred} focus on language-conditioned manipulation tasks that require following sequences of natural-language instructions, and LIBERO~\cite{liu2023libero} extends the paradigm to lifelong learning, benchmarking continual adaptation and knowledge transfer across sequential tasks. Taken together, these frameworks trace a progression from early feasibility to scalable, multimodal, and long-horizon evaluation, enabling large-scale experimentation under controlled conditions.


\subsection{Challenges and gaps}
\begin{figure} 
	\centering
	\includegraphics[width=0.8\textwidth]{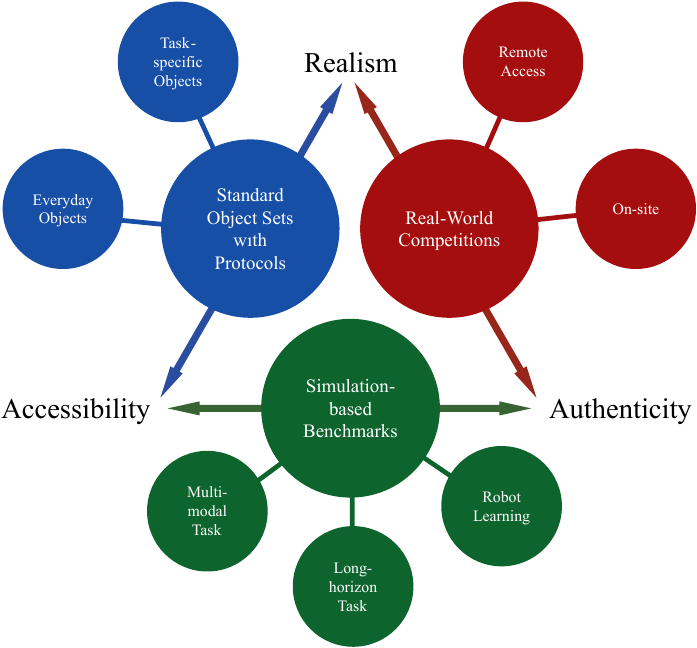} 

	\caption{\textbf{The ``impossible trinity" of the existing robotic manipulation benchmarking effort.} 
		Prior initiatives can be broadly grouped into three categories: standardized object sets with task protocols, real-world competitions, and simulation-based benchmarks. Each category, however, addresses at most two of the required aspects in theory, and none successfully balances realism, accessibility, and authenticity to function as a large-scale, real-world manipulation benchmark.}
	\label{fig:survey} 
\end{figure} 

Although previous benchmarking efforts have advanced the field, they also reveal recurring limitations. Standardized object sets provide a common foundation for reproducible experiments, and when paired with task protocols they establish clear definitions of what should be attempted. However, reproducibility on the task setup itself does not ensure the task authenticity, which explicitly requires verifying the identical adherence to the task protocol and the integrity of the reported results. In practice, when benchmarks are presented solely in papers or videos, the community has little assurance that the protocols were strictly followed and the results were strictly evaluated under the same standard. Without a mechanism for formal evaluation, results obtained with standardized objects remain vulnerable to limited comparability.

In contrast, real-world competitions ensure task authenticity and provides comparable results, as all systems are evaluated under identical conditions and their results are directly observable by the community. However, such events are resource-intensive and require substantial investment in hardware, travel, and logistics, which inevitably excludes many potential participants. Competitions are also geographically centralized or temporally constrained, restricting involvement to those able to attend a single venue or at a fixed period of time. Moreover, their one-off nature makes it difficult to replicate the experimental conditions, limiting reproducibility over time after the event concludes. As a result, while competitions are valuable, they fall short of offering the accessibility and continuity needed for large-scale, inclusive benchmarking.

To address concerns of accessibility and authenticity, simulation-based benchmarks enable scalable experimentation with controllable conditions in the digital world at low cost. However, their reliance on simplified models and physics engines limits realism, preventing them from capturing the full complexity and variability of real-world manipulation. Sensor signals in simulation are likewise idealized, lacking the noise, drift, and calibration challenges inherent to physical systems. Even large-scale scenarios cannot replicate the open-ended diversity of contact uncertainties and interaction dynamics encountered in unstructured environments. \textit{Manipulating atoms is more than manipulating bits}: strong performance in simulation often overestimates a system’s capacity to operate reliably in the wild. Thus, while simulation benchmarks enable massive scalability, they lack the realism required for a trusted measure of genuine manipulation capability for real-world deployment.

In summary, existing approaches each advance robotic manipulation yet face persistent challenges: object sets risk questionable authenticity without formal evaluation, competitions lack accessibility for broad participation, and simulations sacrifice realism for scalability. As summarized in Figure~\ref{fig:survey}, these limitations underscore the need for new paradigm designs that reconcile realism, accessibility, and authenticity while remaining scalable to the demands of global research.
	

\subsection{Rethinking the paradigm of manipulation benchmarking}
The challenges of existing benchmarks suggest that the future of benchmarking manipulation lies not in privileging one category over another, but in designing a paradigm that balances the need for centralization with the benefits of decentralization. Centralization is indispensable for ensuring authenticity, as seen in competitions where tasks are strictly controlled and results directly measured and observed. Meanwhile, decentralization enables accessibility, allowing researchers worldwide to benchmark flexibly within their own environments and robotic system at their convenience without prohibitive costs or logistics. At the same time, realism remains equally indispensable: benchmarks must faithfully reflect the complexity of real-world evaluation if they are to serve as trusted measures of real-world manipulation capability. The challenge, therefore, is to integrate the rigor of centralization with the inclusivity of decentralization while preserving realism.

With this foundation, it is necessary to examine the essence of benchmarking robotic manipulation itself, in order to identify when centralization is indispensable and when decentralization can be selectively enabled. At its core, a manipulation benchmark evaluates a robotic system by observing the process and outcomes of executing well-defined tasks. Since the robotic system is the subject of evaluation, such a process reduces to two major variables: how the task setups are defined and how performance is measured. These two dimensions fundamentally determine the benchmark's authenticity, accessibility, and realism.

Standardized task setups form the foundation of comparable benchmarks. To support large-scale participation, such setups must be reproducible in a decentralized fashion: the same task should be executable by any research group, anywhere, at any time. At the same time, evaluation must balance accessibility with authenticity. While performance data can be collected decentrally, results must be verified centrally against common protocols and metrics to ensure they are trustworthy and comparable. Together, reproducible setups and verifiable evaluations provide the foundation for benchmarking manipulation at scale.

Furthermore, task selection is critical to hierarchically identify the limitations of manipulation capability. Effective benchmarks rely on tasks that are short, well-defined, and designed with stepwise difficulties, so that outcomes are easy to compare and failures are traceable to clear causes. Such tasks provide high diagnostic value while enabling comparability across diverse systems. Once fundamental skills and abilities are reliably benchmarked, they can be systematically composed into more complex, long-horizon tasks that reflect real-world conditions. In this layered approach, carefully chosen primitives ground benchmarks in pinpointing capabilities, while their integration extends evaluation toward realism and practical relevance.

Finally, benchmarks must be designed with sufficient depth to remain relevant for a certain period of time. Tasks that are solved too quickly risk becoming obsolete, offering only a snapshot of current capability. By contrast, durable tasks with difficulty ladders provide enduring reference points, simultaneously documenting progress and driving the field forward by setting meaningful long-term objectives.

\subsection{Benchmarking real-world robotic manipulation at scale}

\begin{figure} 
	\centering
	\includegraphics[width=\textwidth]{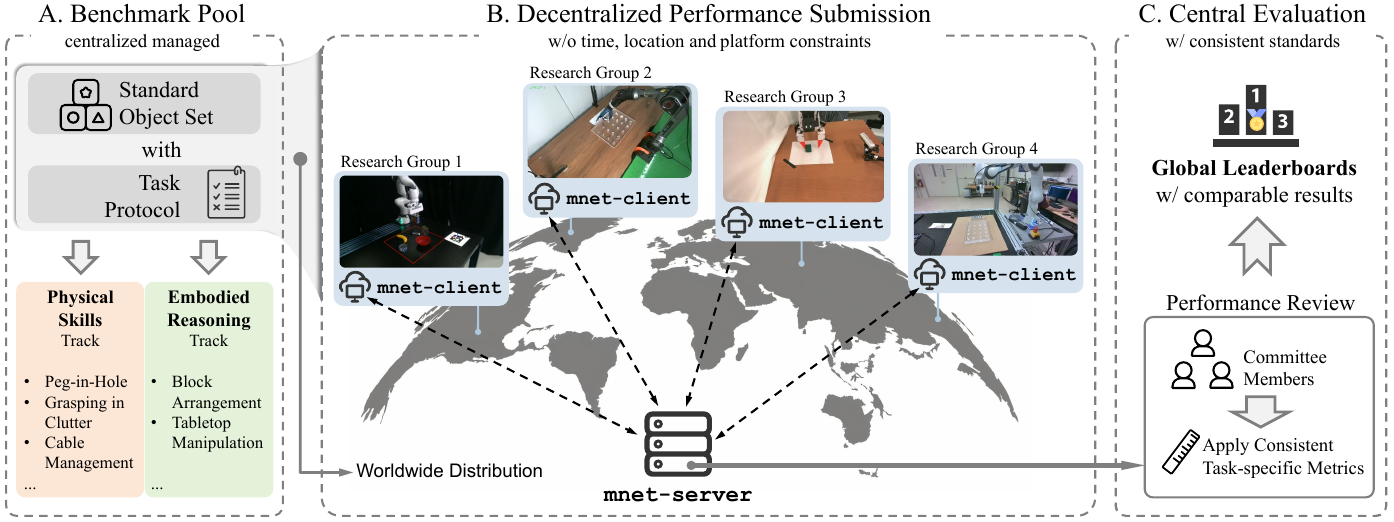} 

	\caption{\textbf{The overall structure of how ManipulationNet operates.}
		Standard object sets are centrally designed and distributed to ensure reproducible task setups. Upon registration, each research group executes the selected task with their customized robotic systems and reports performance through the \texttt{mnet-client} at any time and from any location. All submitted results are centrally evaluated using unified metrics and validated by a human, providing trustworthy and comparable performance assessments across the globe.}
	\label{fig:protocol} 
\end{figure}

Building on these principles, ManipulationNet is an infrastructure designed to host diverse tasks for benchmarking real-world robotic manipulation at scale. The infrastructure stands out from existing paradigms by striking a balance between accessibility, realism, and authenticity. As shown in Figure~\ref{fig:protocol}, this balance is achieved through 1) worldwide distribution of standardized, reproducible task setups, 2) a server–client architecture that enables decentralized performance submission, and 3) centralized verification of results. Importantly, ManipulationNet is governed by a dedicated committee that oversees task selection, resource distribution, and evaluation integrity. Designed to host a broad spectrum of tasks, the infrastructure supports the growth of an interconnected network of manipulation capabilities. The remainder of this section details how ManipulationNet operationalizes principled task standardization and verifiable evaluation at scale.

\textbf{ManipulationNet delivers standardized task setups at scale.} Each task setup is defined through two complementary aspects: a physical object set and a task protocol that governs its use. Prior works~\cite{calli2015ycb, kimble2020benchmarking} have shown that distributing common objects improves comparability across sites while broadening participation. Building on this foundation, ManipulationNet defines object sets as the physical artifacts on which manipulation is evaluated. To ensure long-term availability and consistent quality, these artifacts are centrally designed and distributed globally. Each object set is paired with a protocol specifying initial conditions, manipulation goals, and success criteria. This dual specification evaluates systems not only on identical objects but also under identical task definitions. Crucially, because these setups can be reproduced at any place and any time, they remove the temporal and spatial barriers that have traditionally limited reproducibility. Together, object set with task protocol enables scalability through worldwide distribution and provide diagnostic function for comparable research across diverse robotic systems.

\textbf{ManipulationNet evaluates authentic task performance at scale.} While task setups can be expanded effectively through standardized objects and unified protocols, authentic evaluation at scale presents a greater challenge. At its core, task evaluation involves applying well-defined metrics to observed manipulation performance, verifying goal achievement and adherence to protocol. Purely centralized evaluation maximizes authenticity through direct observation, but is limited by time and location. Conversely, purely decentralized evaluation expands participation but raises concerns about inconsistent measurement or selective reporting. To reconcile these trade-offs, ManipulationNet decouples performance collection from result verification: trials are collected decentrally via \texttt{mnet-client}, while final verification occurs centrally. The process is coordinated by an internet-hosted \texttt{mnet-server} connected in real time to distributed \texttt{mnet-clients}. Once a \texttt{mnet-client} is launched, the trial is immediately registered on the \texttt{mnet-server} to preclude entries which only showcase best performances. During execution, task status is continuously logged; upon completion, the \texttt{mnet-client} submits the recorded video and execution metadata. Integrity is enforced by design as videos cannot be pre-recorded or altered, and execution states are bound in real time to the benchmark protocol. Submitted results are then audited by the ManipulationNet committee, ensuring benchmarked performances remain trustworthy, reproducible, and comparable across the community. In this way, ManipulationNet establishes a standardized evaluation mechanism that combines decentralized reporting with centralized auditing, balancing authenticity and accessibility while unifying evaluation practices.

\textbf{Beyond individual tasks, ManipulationNet aims to construct a network of abilities and skills.} The framework begins with primitive tasks, defined as short and representative tasks that each require merely a specific type of skill or ability to solve. By avoiding unnecessary heterogeneity and complexity, primitive tasks are intrinsically diagnostic as outcomes are unambiguous, failure modes are traceable, and performance can be assessed with minimal subjectivity. Furthermore, primitive tasks also function as building blocks for scalability. Once reliably benchmarked, they can be composed into long-horizon tasks that more closely reflect the demands of general manipulation in unstructured environments. Drawing inspiration from the NIST Assembly Task Board (ATB)~\cite{kimble2020benchmarking}, the initial release of ManipulationNet will emphasize assembly-oriented skills including peg-in-hole, threading and fastening, belt routing and tensioning, and cable management. Over time, the framework will expand to include increasingly diverse tasks, advancing toward comprehensive coverage of general robotic manipulation.


In summary, ManipulationNet establishes a unified benchmark framework that integrates standardized task setups, scalable and authentic evaluation, and a structured progression to cover diverse physical skills and cognitive abilities. By decoupling performance collection from central verification and anchoring all tasks in reproducible object–protocol pairs, it ensures both accessibility and comparability across diverse robotic systems.


\section{Result}
In the following section, we describe the implementation details of ManipulationNet, translating its high-level design into a practical benchmarking infrastructure. All tasks hosted within the framework are executed under a common submission protocol, which relies on the server-client infrastructure. We first introduce this submission protocol, which combines the usage of object sets and \texttt{mnet-client} to enable comparable assessments of real-world manipulation tasks. Next, we detail the server–client mechanism, which provides accessibility while minimizing dependence on network conditions and ensuring result integrity. Finally, we exemplify the Physical Skills Track and the Embodied Reasoning Track with benchmark tasks included for the first release.

\subsection{Benchmarking Protocol}

\begin{figure} 
	\centering
	\includegraphics[width=\textwidth]{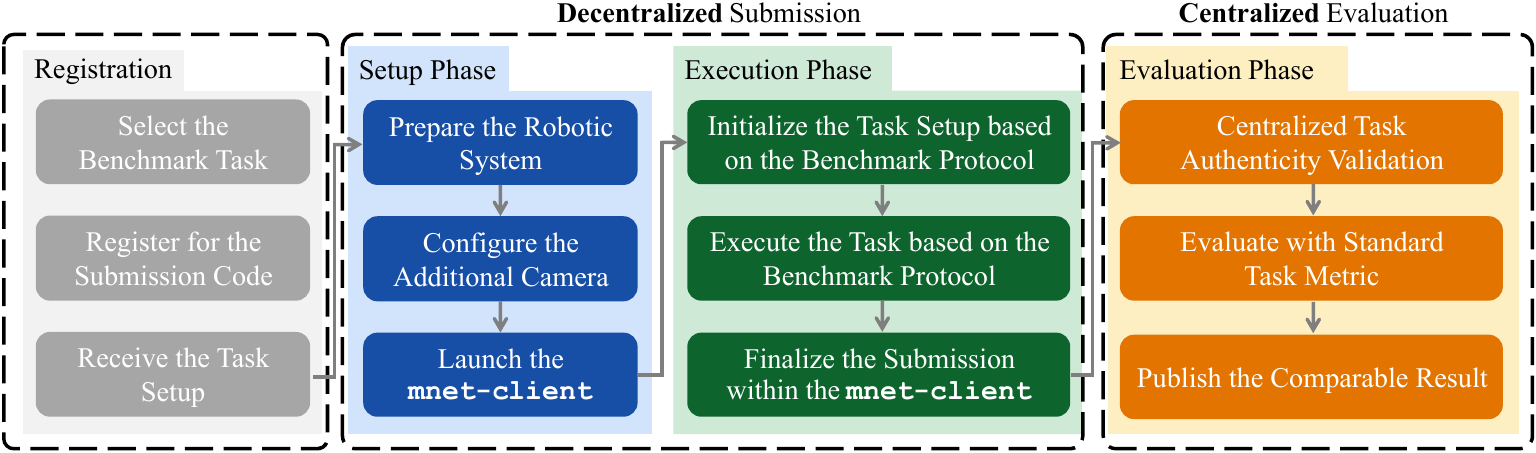} 

	\caption{\textbf{Overview of the general protocol that applies to all hosted tasks.} Once the research group receives the physical object set, they can reproducibly configure task setups locally with their customized robotic system. To formally benchmark the system performance, the only additional requirement is connecting an external camera to the \texttt{mnet-client} and launching it prior to the task execution. Performance data are then transmitted to the \texttt{mnet-server}, where they are evaluated by official judges according to unified metrics, ensuring comparability across submissions. Under this protocol, manipulation performance is collected in a decentralized manner through distributed \texttt{mnet-client}, while final verification is conducted centrally.}
	\label{fig:example} 
\end{figure}

In this subsection, we present the general benchmarking protocol that governs all tasks hosted by ManipulationNet as shown in Figure~\ref{fig:example}. This general protocol is a standalone high-level procedure designed for performance reporting and task instruction delivering. For the setup phase, participants first receive the standardized object set and configure their robotic system according to the benchmarking task. To ensure unbiased observation, an independent camera, separate from the robot hardware, is connected to the \texttt{mnet-client} to record the execution process. Once the \texttt{mnet-client} is launched, the trial is immediately registered on the \texttt{mnet-server}, and each participant is allocated a limited number of trials within a given period to prevent selection bias through repeated attempts.

During the execution phase, the \texttt{mnet-server} first generates a random one-time submission code and transmits it to the \texttt{mnet-client}. Participants must display this code within the camera’s field of view, uniquely binding the recording to the session that ensuring all events occur after the \texttt{mnet-client} initialization. From that point onward, the \texttt{mnet-client} and \texttt{mnet-server} maintain a secure, stable connection through which 1) the \texttt{mnet-client} reports task execution status to the \texttt{mnet-server} in real time, and 2) the \texttt{mnet-server} delivers task instructions to the \texttt{mnet-client}, this could involves language/visual prompts, task-specific instructions, and more. Upon completion of tasks execution, the \texttt{mnet-client} transmits the recorded video and execution logs to the \texttt{mnet-server}.

Submitted performances are verified centrally by judges from the ManipulationNet committee. As detailed in the following subsection, server–client mechanisms guarantee the integrity of submitted videos and execution logs. Centralized evaluation then applies task-specific metrics consistently across all submissions to ensure objectivity. Finally, verified results are published on the official ManipulationNet platform with participant consent, establishing a transparent and comparable record of benchmark outcomes for the community.

\subsection{Server-Client Mechanism}
To balance decentralized participation with centralized trust, ManipulationNet employs a server–client mechanism that ensures both accessibility and verifiable integrity in task submissions. The implementation of the server-client mechanism is detailed in Section~\ref {server-client}. Each submission includes a complete video covering task initialization, execution, and completion, supplemented by real-time execution logs and selected key frames transmitted to the \texttt{mnet-server}. To minimize bandwidth requirements, the \texttt{mnet-client} does not stream raw video during execution. Instead, it transmits only lightweight metadata, such as task events, status messages, and cryptographic hashes of video frames, ensuring accessibility even under constrained network conditions.

To guarantee authenticity for each trial, in addition to requiring that the one-time submission code remain visible in the video, the \texttt{mnet-server} issues random requests during task execution. For each request, the \texttt{mnet-client} extracts the corresponding video frame locally, computes its hash and transmits the hash value to the \texttt{mnet-server} in real time.  In parallel, the \texttt{mnet-client} is required to report the task execution status in real-time to the \texttt{mnet-server} to enable future cross-check on video content against reported task events at identical timestamps. Communication between the robotic system and the \texttt{mnet-client} is enable through Robot Operation System (ROS) services and topics. At task completion, the \texttt{mnet-client} computes a final hash of the entire video, transmits the hash to the \texttt{mnet-server} for completeness verification, and only then compresses and uploads the complete submission package, including the complete video, selected frames, and metadata. Because all integrity-relevant data are already logged in real time, the upload may take as long as necessary without compromising guarantees, even under poor network conditions.

Once a submission is received by the \texttt{mnet-server}, the ManipulationNet committee validates it against three criteria: 1) the recorded video must clearly display the one-time submission code, 2) the uploaded video and frames must match the hash values previously registered with the \texttt{mnet-server}, and 3) the video length and content must align with the reported task status and key frames at each corresponding timestamp. The hash function is highly sensitive to even a one-byte change, so if the uploaded files produce the same hash value, the files can be verified unaltered. Task-specific performance metrics are applied only after these integrity checks have been passed, ensuring objective and comparable evaluations across systems.

Through this mechanism, ManipulationNet ensures that every performance submission is uniquely tied to its registered trial, manipulating under a standardized task setup, resistant to both pre-recording and post-modification, and fully verifiable by the committee. The result is a secure, bandwidth-efficient protocol that balances broad accessibility with rigorous integrity guarantees.

\subsection{Physical Skills Track}
\label{pih_sec}

The Physical Skills Track benchmarks fundamental interaction skills in manipulation, focusing on how robots adaptively and purposely achieve the specified goal under physical constraints. In this subsection, we present the hosted tasks under the physical skills track for the first release, which includes peg-in-hole assembly, cable management and grasping in clutter.

\subsubsection{Peg-in-Hole Assembly}
\begin{figure}[h] 
	\centering
	\includegraphics[width=\textwidth]{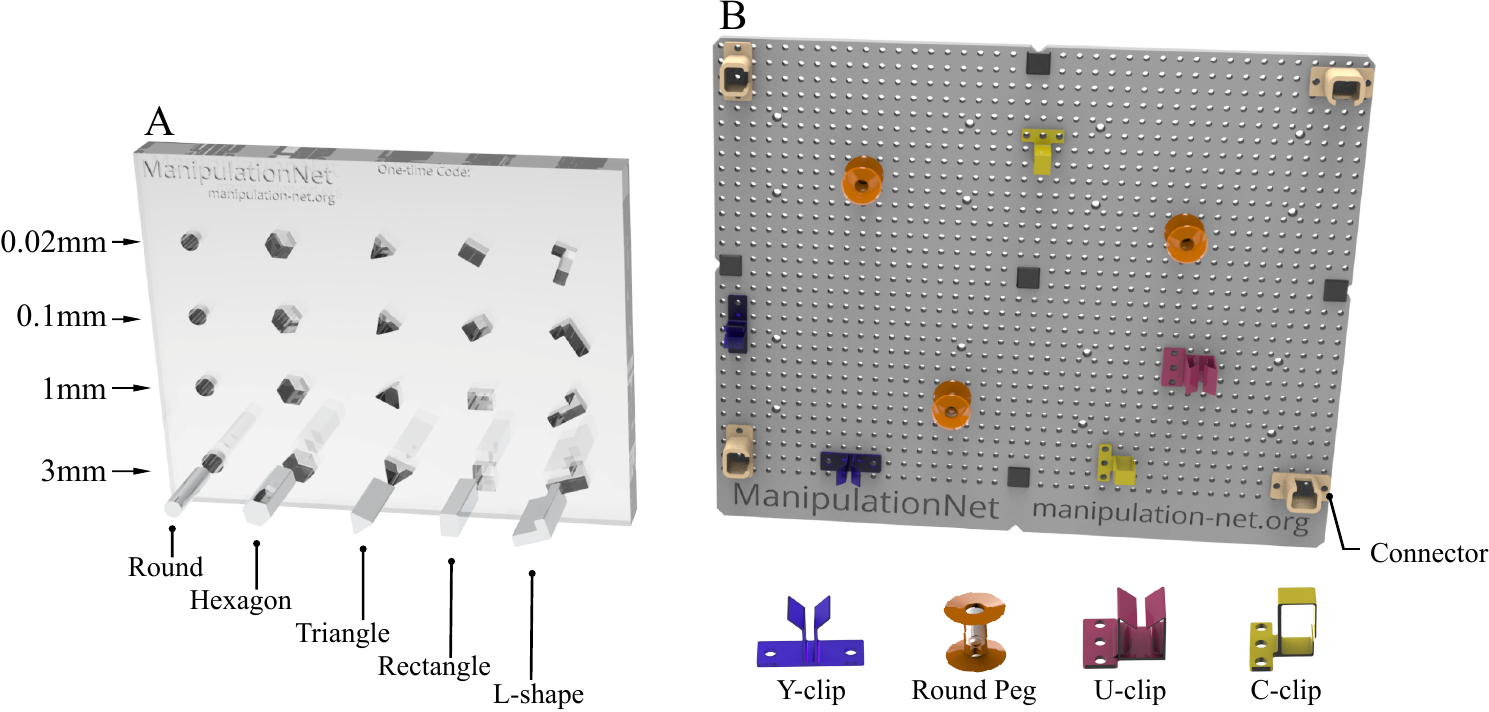} 

	\caption{\textbf{The design of the peg-in-hole assembly and cable management benchmarks.} A. The board design comprises five distinct shapes, encompassing both symmetric and asymmetric forms. For each shape, four clearance levels were implemented, with tolerances of 0.02 mm, 0.1 mm, 1 mm, and 3 mm. The assembly board is fabricated from transparent acrylic, with a manufacturing tolerance within 20 microns, to introduce perceptual and physical challenges, while the pegs were constructed from stainless steel to ensure robustness and durability. B. Four different types of fixtures, including the Y-clip, Round Peg, U-clip, and the C-clip, are involved to flexibly define a wide range of cable routing configurations. Since cable routing does not require high precision, the board and the fixture are distributed through 3D printing files. }
	\label{fig:mnet_pih} 
\end{figure}

The peg-in-hole assembly benchmark demands robust adaptability and generalization under real-world contact dynamics to achieve high-precision goals. The assembly pegs and board, as the object set, are designed to systematically vary both geometry and clearance, thereby enabling comprehensive evaluation of a system’s ability to generalize across increasing levels of insertion difficulty. As illustrated in Figure~\ref{fig:mnet_pih}-A, the peg set includes five distinct shapes, ranging from highly symmetric to highly asymmetric geometries. For each shape, the corresponding hole is fabricated in four clearance levels, with tolerances of 3 mm, 1 mm, 0.1 mm, 0.02 mm, spanning from loose to extremely tight fits. To further raise perceptual difficulty and provide a clear view of the insertion process, the task board is manufactured from transparent acrylic material. Transparent material poses a well-known challenge for vision systems, and we hypothesize that systems capable of handling them robustly will generalize more readily to objects with varied colors and textures. Since the peg-in-hole task only requires task execution in a fixed order, the server only collects execution status in real-time during the benchmarking process. Together, these design elements provide reproducibility while testing both perceptual and physical robustness. 

\subsubsection{Cable Management}
The cable management benchmark requires robustly manipulating deformable linear objects to achieve specified routing configurations under environmental contact constraints. The cable, fixtures, and board, as the object set, are designed to define a wide range of routing configurations, allowing for a comprehensive evaluation of the robot's long-horizon planning and interaction skills. As illustrated in Figure~\ref{fig:mnet_pih}-B, four types of fixtures, including C-clip, round peg, U-clip, and Y-clip, are adopted to hold the cable during the routing process. Since this task does not require high precision for hardware manufacturing, all the fixtures and boards are distributed through 3D printing files. Fixtures can be flexibly mounted on the holes in the task board to provide various contact constraints. During benchmarking, the server will send the target routing configuration to the client in real-time, and the participant team is asked to reproduce it accordingly for system evaluation. Together, these design elements provide reproducibility while allowing for a wide range of variation, enabling the generalizable evaluation of cable management.

\subsubsection{Grasping in Clutter} \label{benchmark_grasp_in_clutter}
\begin{figure}
    \centering
    \includegraphics[width=\linewidth]{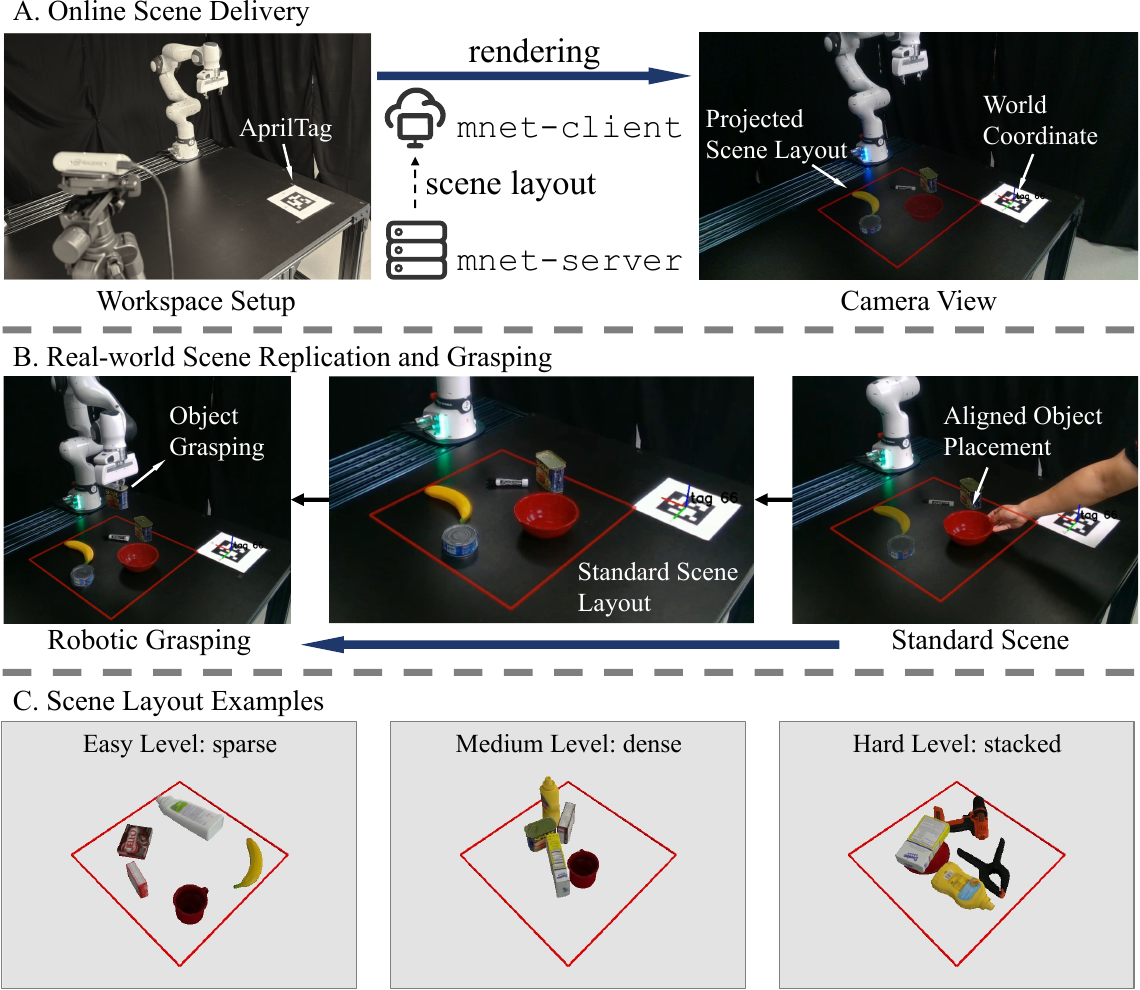}
    \caption{\textbf{The overview of the Grasping in Clutter task.} (A) An AprilTag is used to establish the world coordinate frame in the real environment, enabling the server to render the corresponding projected scene layout. (B) After the mnet-client receives this projected scene, a human operator places all objects according to the projection mask to construct the standardized grasping testbed. (C) The benchmark includes three difficulty levels: easy scenes with sparsely arranged objects, medium scenes with objects positioned in close proximity, and hard scenes featuring stacked objects.}
    \label{fig:grasping}
\end{figure}
The grasping in clutter benchmark requires the robotic system to reliably grasp a diverse set of objects from the tabletop. This benchmark provides standard tabletop scene layouts based on the concept of SceneReplica~\cite{khargonkar2024scenereplica} and utilizes 16 objects from the YCB object set to achieve comparable manipulation results. As shown in Figure~\ref{fig:grasping}, after placing a standard AprilTag~\cite{olson2011apriltag} on the tabletop surface to introduce a virtual world coordinate. During benchmarking, the \texttt{mnet-client} will receive a scene layout from the \texttt{mnet-server} in real-time and further project its corresponding scene mask that overlays with the real-world environment. The human operator is required to place all the objects involved on the tabletop to align with the projected scene for the standard scenario setup. Each scene layout contains 5 objects, and the robot is required to grasp each object one by one and place it inside a container (e.g., a box) outside the projected red square area. The scene layouts are characterized by varying object densities, including sparse, dense, and stacked configurations. The grasping in clutter benchmark evaluates the grasping performance based on the declutter rate (the ratio of objects successfully grasped and removed to all spawned objects), the grasp success rate (the ratio of successful grasps to the total number of grasp attempts), and time efficiency.

\subsection{Embodied Reasoning Track}
\label{vla_sec}

The Embodied Reasoning Track evaluates manipulation from a reasoning and multimodal interpretation perspective, focusing on how robots integrate language and visual perception into grounded physical actions. In contrast to the Physical Skills Track, which imposes strict demands on physical interaction, this track deliberately lowers physical difficulty to isolate and diagnose reasoning failures when they occur. In this subsection, we present the hosted tasks under the Embodied Reasoning Track for the first release, which includes block arrangement and tabletop management.

\subsubsection{Language-conditioned Tabletop Manipulation}
The language-conditioned tabletop manipulation benchmark assesses how robotic systems take natural language instructions as explicit conditioning signals for tabletop object manipulation in the real world. During benchmarking, the human operator is required to set up the initial scene layout for each task within the task region, following the same procedure as the grasping in clutter benchmark~\ref{benchmark_grasp_in_clutter}. Additionally, the \texttt{mnet-server} will also send an accompanying language instruction to the \texttt{mnet-client} with each scene layout.  The manipulation system is required to follow the given language instruction to rearrange the tabletop scene to achieve the specified goal. 
As shown in Figure~\ref{fig:mnet_vla}-A,B, diverse physical skills are included for a comprehensive evaluation. The natural language instruction would primarily involve understanding object instances and spatial relations between them. The object instance would be directly or indirectly described by its name, color, category, and functionality. This benchmark evaluates a system's ability to execute language-conditioned tabletop manipulation that requires clear instance-level understanding and spatial reasoning.

\subsubsection{Block Arrangement}
\begin{figure}[h] 
	\centering
	\includegraphics[width=\textwidth]{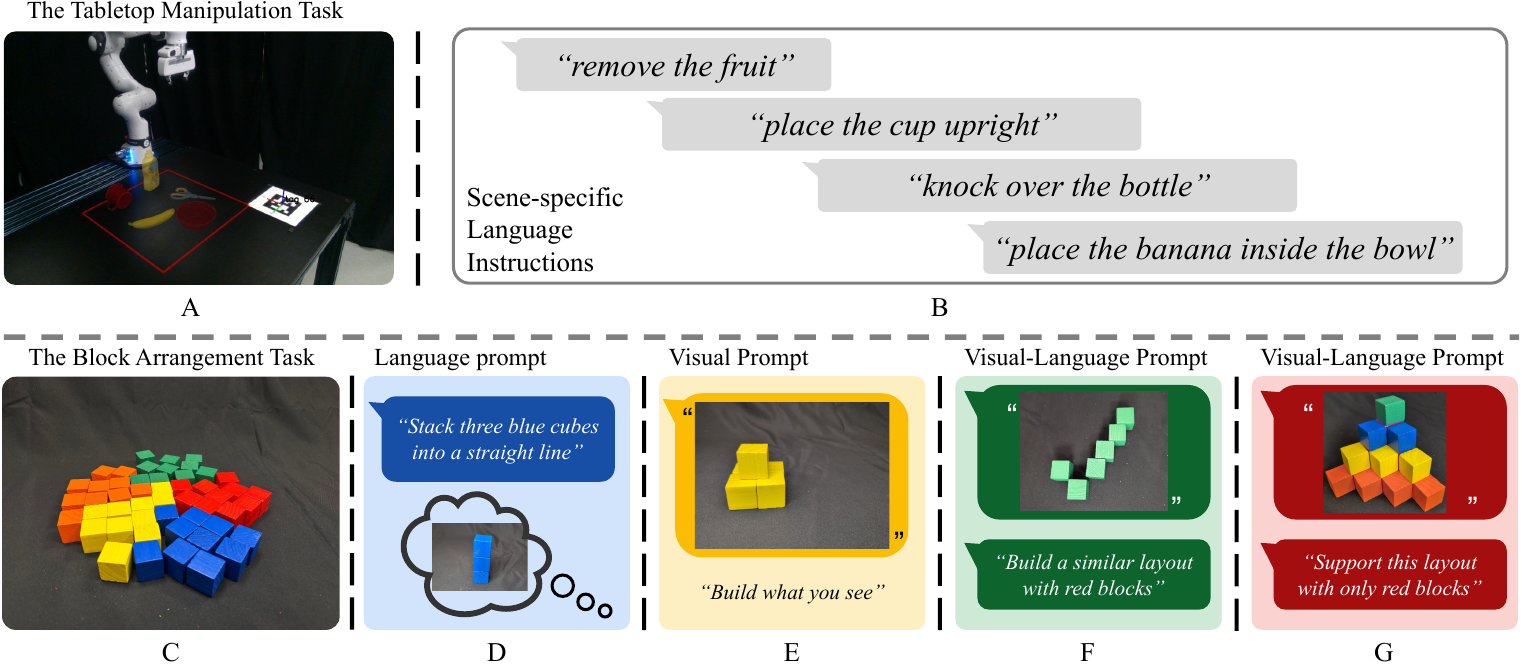} 

	\caption{\textbf{The overview of the tabletop manipulation and the block arrangement benchmarks.} A. The projected scene layout for the language-conditioned tabletop manipulation through \texttt{mnet-client}; B. The example language instructions for the projected scene layout; C. The object set for the block arrangement task. In this task, the robot is required to plan and execute manipulation actions to arrange the blocks according to instructions received from the \texttt{mnet-server}. D. An example task with a language prompt. E. An example task with a visual prompt, where the default requirement is to replicate an observed real-world arrangement. F. An example of a visual-language prompt without visual occlusion. G. An example of a visual-language prompt that incorporates visual occlusion, requiring physics and spatial reasoning about support structures to maintain the layout while also satisfying the color constraints.
	  }
	\label{fig:mnet_vla} 
\end{figure}

The block arrangement benchmark tests how robotic systems translate natural language and visual instructions into executable actions to achieve the specified goals. As shown in the Figure~\ref{fig:mnet_vla}-C, the object set comprises colored blocks in five distinct colors \{red, yellow, orange, blue, green\}, with ten blocks per color. All blocks are identical in shape, size, and material. During benchmarking, the manipulation system is required to reproduce the block layout in the real world, as specified by the received prompts. This process may involve simple actions to manipulate the blocks, such as pick-and-place. The target capability for evaluation is to interpret 1) language prompts; 2) visual prompts and 3) visual-language prompts into a specified goal and its associated composition of actions.
For example, in the language prompt, simpler instructions may require forming a straight line of blocks, whereas more advanced instructions may demand three-dimensional structures: \textit{“Stack three blue cubes into a straight line”} as shown in Figure~\ref{fig:mnet_vla}-D. For the visual prompt, the robot is asked to build what it sees from the image, even if it is only a partial view, as exemplified in Figure~\ref{fig:mnet_vla}-E. The robot needs to reason about its possible support structure to enable a physically stable layout in the real world. For visual-language prompts, the task cannot be solved by relying on either vision or language alone, as indicated in Figure~\ref{fig:mnet_vla}-F, G. The robot must integrate information from both modalities to infer the correct block layout and, consequently, the appropriate sequence of actions. As a diagnostic benchmark, it measures a system’s capacity to interpret and act upon cross-modal instructions, while minimizing confounds from contact-rich dynamics.

 \subsection{Baseline Results}
 \begin{figure}
     \centering
     \includegraphics[width=\linewidth]{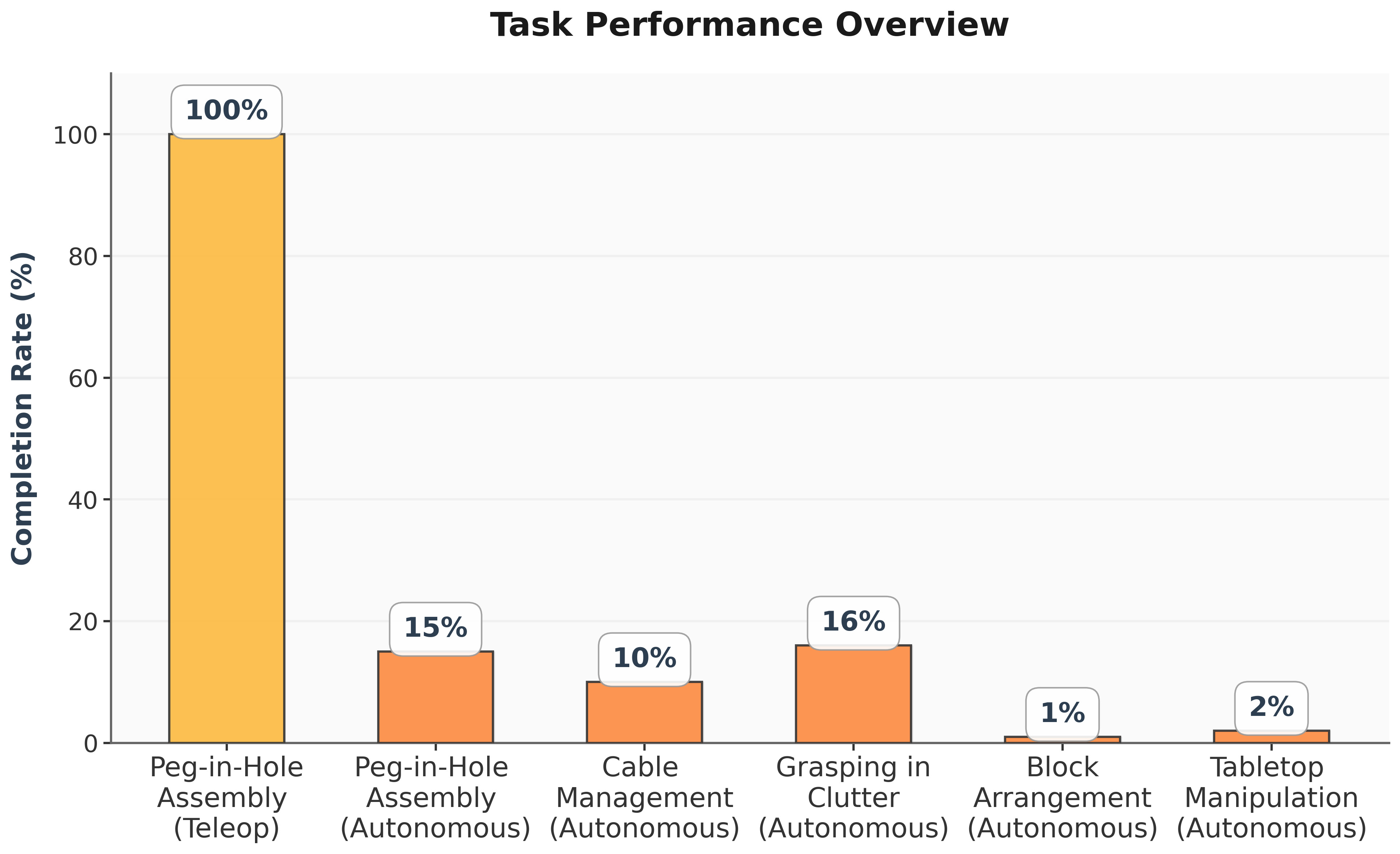}
     \caption{\textbf{The summary of the baseline results for each benchmark.} The score for each task is normalized as a percentage of its maximum possible score to provide an intuitive measure of progress. Statistics presented in this figure is preliminary results obtained during the test period of ManipulationNet. Data will be updated as we receive more submissions.}
     \label{fig:output}
 \end{figure}
 
 Given the large variations across tasks and the fact that most existing algorithms are tailored for a single functionality, the baseline result for each benchmark is obtained with task-specific methods. All experiments are carried out in the real world which strictly follows the benchmarking protocol defined by each specific benchmark. For more details, please refer to the supplementary material as hosted on \url{manipulation-net.org}.

 The summary of the preliminary result is visualized in Figure~\ref{fig:output}. The score for each task is normalized as a percentage of its maximum possible score to provide an intuitive measure of progress. For more details regarding the implementation of each manipulation system, please refer to the supplementary material.

\section{Discussion}
ManipulationNet aims to establish a durable, transparent, and community-driven framework for benchmarking robotic manipulation at scale. While the technical details presented in this paper describe the design and implementation of its release version, the ultimate goal is not merely to host a set of tasks, but to guide the field toward a systematic understanding of robotic manipulation capability over time. To this end, we outline the mission of ManipulationNet across different timeframes, from immediate goals to future research directions.

In the near term, our priority is to unify the community around a small set of well-defined benchmark tasks. By lowering barriers to participation and providing standardized protocols and evaluation mechanisms, ManipulationNet aims to encourage broad engagement from diverse research groups. The goal in this phase is not comprehensive coverage, but rather the establishment of a shared platform where results are directly comparable, fostering collaboration and accelerating progress on representative diagnostic tasks.

Over the next several years, we envision ManipulationNet expanding into a richer suite of tasks that collectively capture the breadth of manipulation challenges faced by the community. The goal is to benchmark not only what robots can do, but also how and why they succeed or fail across different task categories. By curating a continually updated set of tasks aligned with both emerging capabilities and persistent challenges, the framework will provide diagnostic depth while also serving as a guidepost for research priorities.

In the longer horizon, ManipulationNet aspires to serve as a historical record of robotic manipulation capability. At this stage, it will address two fundamental questions: \textit{What is robot manipulation capable of at a given point in time?} and \textit{Which capabilities are sufficiently mature for real-world deployment?} Anchored in a standardized yet evolving framework, ManipulationNet will not only document the trajectory of scientific progress but also help bridge the enduring gap between laboratory demonstrations and practical adoption.

Ultimately, the vision of ManipulationNet is to evolve in concert with the field itself: beginning with a handful of tasks to unify effort, scaling to a broad set of benchmarks to guide inquiry, and culminating in a durable platform that both records progress and informs real-world readiness. Through this progression, we hope to establish ManipulationNet as a cornerstone of robotic manipulation research, providing a collective foundation for both scientific discovery and technological impact.

\section{Materials and Methods}\label{server-client}

In addition to describing the design philosophy, we next detail the technical implementation of the \texttt{mnet-client} and \texttt{mnet-server}. As a global infrastructure, we host the \texttt{mnet-server} on Amazon Web Services (AWS)~\cite{aws}, for both computing and storage, to support distributed \texttt{mnet-clients} worldwide to report their task performance in a reliable and inclusive way. Communication occurs over the Transmission Control Protocol (TCP) across the public internet, ensuring stable and standardized connections.

\begin{figure} 
	\centering
	\includegraphics[width=\textwidth]{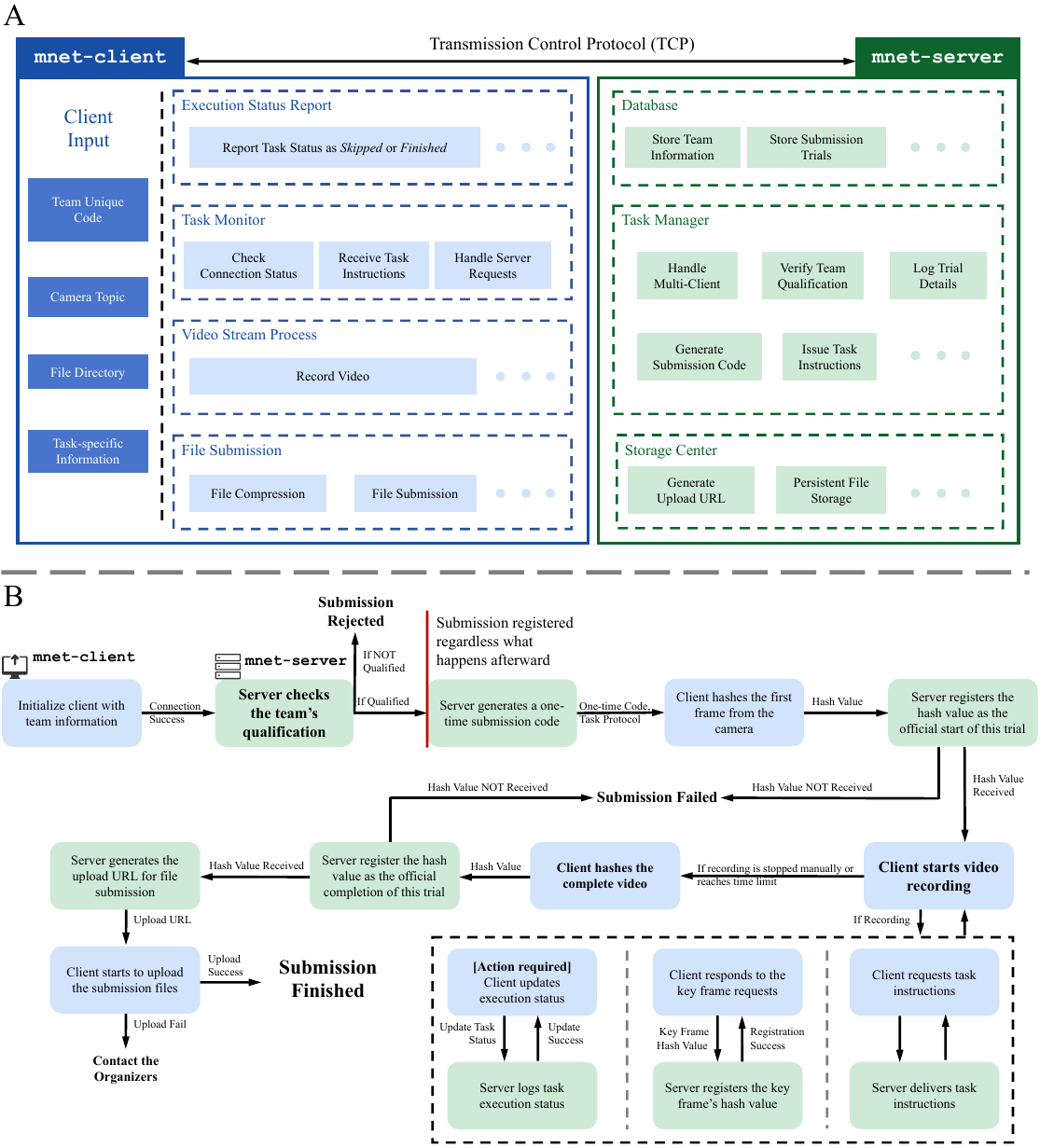} 

	\caption{\textbf{ Detailed protocol of the server-client performance submission. } A. The overview of the software structure of the \texttt{mnet-client} and the \texttt{mnet-server}. B. The overview of the detailed activities for each performance submission.
		}
	\label{fig:framework_protocol} 
\end{figure}

The overview of the software structure is demonstrated in Figure~\ref{fig:framework_protocol}-A. The \texttt{mnet-client} is implemented as a ROS-compatible package so that it integrates naturally with robotic platforms. It requires the registration and task information as input to start the submission trial, which manages execution status reporting, task monitoring, video processing, and file submission. For the execution status report, the \texttt{mnet-client} initializes two ROS services to declare whether the current ongoing task has been finished or skipped. Each service can be called with standard message \texttt{Trigger} and is acknowledged by the \texttt{mnet-server} to confirm receipt. Task monitoring is achieved through regular communication with the \texttt{mnet-server}, during which the \texttt{mnet-client} receives task instructions, responds to keyframe requests, and checks connection status. Relevant messages are further published on ROS topics, ensuring that both robots or humans can access them transparently within the ROS environment. For video recording, the \texttt{mnet-client} subscribes to an independent ROS camera topic and saves the execution process using \texttt{OpenCV} under the specified frame rate. To standardize format and maintain compatibility across diverse playback environments, videos are encoded using the \texttt{x264 codec}. Once execution is complete, the \texttt{mnet-client} compresses the submission package and requests a pre-signed upload URL from the \texttt{mnet-server}. Using this temporary address, it uploads the video, key frames, and metadata via \texttt{HTTP PUT}, thereby minimizing network overhead and ensuring reliability.

On the \texttt{mnet-server} side, each \texttt{mnet-client} connection initiates a dedicated thread to manage all activities (as visualized in Figure~\ref{fig:framework_protocol}-B for that trial and closes once submission concludes. The \texttt{mnet-server}'s architecture consists of a task manager coupled with a storage center. The task manager oversees all message-level operations, including team qualification checks, issuance of one-time submission codes, logging of execution status, and delivery of task-specific instructions. Because transferring large files (up to dozens of GBs) reliably across the globe is challenging, the storage center is implemented using AWS S3. When a submission package is ready, the task manager requests a pre-signed upload URL from S3, valid for two hours, and sends it to the \texttt{mnet-client}. The \texttt{mnet-client} then transfers its data directly to the data center. If an upload fails because of connectivity issues or file size, integrity-critical artifacts such as logs, hash values, and timestamps have already been stored on the \texttt{mnet-server}, allowing the research team to resume uploading later with a new URL without compromising verification. All trial metadata, team records, and submission states are maintained in a MySQL database, ensuring reliable coordination of decentralized submissions with centralized auditing.

Through this architecture, ManipulationNet supports benchmark tasks to be executed on any ROS-enabled robotic system with minimal integration effort, while preserving integrity-ensured video capture, consistent task reporting, and online instruction delivery with the \texttt{mnet-server}.

\clearpage 

%
\bibliography{science_template} 
\bibliographystyle{sciencemag}

\end{document}